\documentclass{article}

\usepackage{arxiv}

\usepackage[utf8]{inputenc} 
\usepackage[T1]{fontenc}    
\usepackage{hyperref}       
\usepackage{url}            
\usepackage{booktabs}       
\usepackage{amsfonts}       
\usepackage{microtype}      
\usepackage{graphicx}

\usepackage{tabularx}
\usepackage[table]{xcolor}
\usepackage{pifont}
\usepackage{array}
\usepackage{geometry}
\usepackage{longtable}

\newcommand{\cmark}{\textcolor{blue}{\ding{51}}}  
\newcommand{\xmark}{\textcolor{red}{\ding{55}}}

\author{
 Seshu Tirupathi\\
  IBM Research Europe\\
  Dublin, Ireland\\
  \And
   Dhaval Salwala\\
  IBM Research Europe\\
  Dublin, Ireland\\
  \And
  Elizabeth Daly\\
  IBM Research Europe\\
  Dublin, Ireland\\
  \And
  Inge Vejsbjerg\\
  IBM Research Europe\\
  Dublin, Ireland\\
}

\date{{May 2014}} 

\begin{document}

\title{GAF-Guard: An Agentic Framework for Risk Management and Governance in Large Language Models}

\maketitle

\begin{abstract}
As Large Language Models (LLMs) continue to be increasingly applied across various domains, their widespread adoption necessitates rigorous monitoring to prevent unintended negative consequences and ensure robustness. Furthermore, LLMs must be designed to align with human values, like preventing harmful content and ensuring responsible usage. The current automated systems and solutions for monitoring LLMs in production are primarily centered on LLM-specific concerns like hallucination etc, with little consideration given to the requirements of specific use-cases and user preferences. This paper introduces GAF-Guard, a novel agentic framework for LLM governance that places the user, the use-case, and the model itself at the center. The framework is designed to detect and monitor risks associated with the deployment of LLM based applications. The approach models autonomous agents that identify risks, activate risk detection tools,  within specific use-cases and facilitate continuous monitoring and reporting to enhance AI safety, and user expectations. The code is available at \url{https://github.com/IBM/risk-atlas-nexus-demos/tree/main/gaf-guard}. 
\end{abstract}

\section{Introduction}
The governance of Large Language Models (LLMs) poses a significant challenge due to the scale and their rapid development. Traditional risks associated with AI models, such as data bias and model interpretability, are compounded by new concerns specific to LLMs, including misinformation and copyright infringement \cite{fang2024large}. 

The generic nature of LLMs makes it difficult to isolate risk assessments solely from an LLM perspective. Instead, a more nuanced approach is needed to consider the context in which these models are used. As noted in \cite{fang2024large}, the clarity of the use-case plays a critical role in defining the risks and evaluating the performance of an LLM.

To illustrate this point, two examples are cited in \cite{fang2024large}: the deployment of LLMs in law enforcement versus educational settings, and internal productivity chatbots versus those directly interacting with patients in healthcare. By examining these differences, we can better understand how to mitigate risks associated with LLMs and ensure their safe and responsible use.

To further drive home this point, we cite three independent studies conducted by McKinsey \cite{mckinsey}, Mindforge  consortium \cite{mindforge}, and UK Finance along with Accenture \cite{ukfinance}. Each of these studies investigated the risks associated with Generative AI broadly in the context of software development lifecycle use-cases.

A summary of the identified risks is presented in Table \ref{tab:risk_comparison}. There is a significant overlap in the risks identified by the three studies with some differences as well where the end-user is able to provide the required nuances for the relevant use-case. This level of granularity and use-case relevance underscores the importance of considering not just the overall set of risks associated with Generative AI, but also the unique contextual factors (like user and use-case) that can influence its adoption and use.

\begin{table}
    \centering
     \setlength{\leftmargini}{0.4cm}
    \begin{tabular}{| m{4cm} | m{4cm} | m{4cm} |}
        \hline
         McKinsey & Mindforge & UK Finance \\
        \hline
        \begin{itemize} 
            \item IP infringement 
            \item Malicious use
            \item Security threats
            \item Explainability
            \item 3rd party accountability
            \item Data leakage
            \item Performance
        \end{itemize} & 
        \begin{itemize} 
            \item IP infringement 
            \item Malicious use
            \item Security threats
            \item Explainability
            \item 3rd party accountability
            \item Data leakage
            \item Performance
            \item Hallucination
            \item Data not fit for purpose
        \end{itemize} & 
        \begin{itemize} 
            \item IP infringement 
            \item Security threats
            \item Explainability
            \item 3rd party accountability
            \item Data leakage
            \item Performance
            \item Environmental impact
        \end{itemize} \\
        \hline
    \end{tabular}
    \label{tab:risk_comparison}
\caption{Risks associated with LLMs for software coding identified by three independent research studies. A lot of common risks are identified with some differences or additional risks mentioned in some studies.}
\end{table}

At the same time, identifying the risks associated with a given use-case can be time consuming and challenging \cite{daly2025usagegovernanceadvisorintent}. Further, even after identifying the risks, LLM governance lifecycle can still be challenging to maintain consistency across different stages of the lifecycle like identifying risks, suggesting mitigation measures, plan mitigation strategies, monitor and audit the model. There are multiple tools that can help with segments of this lifecycle like the examples shown in Table \ref{tab:governance_frameworks}. However, a holistic solution that can assist the user through the lifecycle of LLM governance with focus on the use-case is still missing.

\begin{longtable}{|>{\raggedright\arraybackslash}p{1.4cm}|c|c|c|c|c|c|}
\hline
\textbf{Resource} & \textbf{Implementation} & \textbf{Pre-deployment} & \textbf{Post-deployment} & \textbf{Holistic} & \textbf{User/use-case}  \\
\hline
OWASP \cite{owasp} & \xmark & \cmark & \xmark & \cmark & \xmark  \\
\hline
NIST \cite{nist}  & \xmark & \cmark & \xmark & \cmark & \xmark  \\
\hline
OpenAI/ Evals \cite{openai_evals}    & \cmark & \cmark & \xmark & \cmark & \xmark  \\
\hline
NVIDIA/ Garak \cite{garak}  & \cmark & \cmark & \xmark & \xmark & \xmark  \\
\hline
NVIDIA/ Guardrails \cite{rebedea-etal-2023-nemo}   & \cmark & \xmark & \cmark & \xmark & \cmark  \\
\hline
\textbf{GAF-Guard}   & \cmark & \cmark & \cmark & \cmark & \cmark  \\
\hline
\caption{Comparison of standards, tools, and resources available for governance of LLMs.}
\label{tab:governance_frameworks}
\end{longtable}

To address this gap, we propose GAF-Guard (Governance Agentic Framework), an agentic AI framework that can effectively detect and manage risks associated with LLMs for a given use-case. The framework leverages agents to identify risks tailored to a specific use-case, generate drift and risk monitors using pre-deployment risk assessment, and establish real-time monitoring functions for LLMs post deployment. By integrating these capabilities, our approach aims to provide a 
comprehensive risk management framework that addresses the unique requirements of each LLM application for a given use-case.


This paper is organized as follows: Section \ref{related_work} reviews related work. Section \ref{agent_based_gov_framework} presents the governance framework, including agent descriptions. 
Section \ref{evaluation} describes the dataset, implementation, and evaluation of the framework. Finally, Section \ref{conclusion} provides the summary and future work directions.  

\section{Related Work}
\label{related_work}

The advent of LLMs has significantly fueled the research and the development of a vast array of language-related applications, and demonstrated their capacity to drive innovation and growth across diverse fields (ex. \cite{zeng2025examples}, \cite{neumann2024llm}). 

LLMs have also sparked significant interest in the design and development of agentic systems, which are engineered to operate independently, leveraging a cyclical process of thinking, reasoning, acting, and continuous learning to stay adaptable and effective \cite{putta2024agent}. Agentic applications have seen a notable acceleration in their development, with a corresponding increase in the range of potential use-cases and applications. Example domain applications include healthcare where patient data can be analyzed and alert corresponding medical personnel, analyze and detect fraudulent activities, intelligent tutoring systems etc (Section 5, \cite{acharya2025agentic}). 

However, the successful application of LLMs for inference tasks depends on robust benchmarking metrics, methodologies and effective monitoring strategies. For example, G-eval (\cite{liu2023g}, \cite{Ip_deepeval_2025}) provides an LLM-as-judge framework with Chain-of-Thought (CoT) to measure the quality of texts generated by LLMs and demonstrate that these metrics perform better than conventional metrics like BLEU and ROUGE scores. To address LLM monitoring and governance, a range of observability systems have been developed to track the performance and behavior of LLMs and agentic systems in production environments. Research efforts mentioned in Table \ref{tab:governance_frameworks} are a few examples in this space. While numerous libraries and research works have been developed to address specific aspects of LLM governance and operations, they tend to focus on individual components rather than offering a comprehensive, holistic perspective on governance.

On the agents side, an example of agent monitoring tool is AgentGuard \cite{chen2025agentguard}, an agentic framework that autonomously covers unsafe tool-use workflows and generating safety constraints. Another example of agent monitoring tool is {$\tau$-bench} \cite{yao2024tau} which measures the average performance of the overall agentic system over multiple runs of the system. 

In contrast to the works above, our work presents an integration of agentic systems and monitoring technologies, yielding a novel agentic governance framework to detect and monitor LLM risks with ease throughout the lifecycle of the model.

\section{Agent based Governance framework}
\label{agent_based_gov_framework}
In this section, we describe the architecture of GAF-Guard. First, we describe Risk Atlas Nexus, an open source library, that provides APIs for governance of Foundation Models, which is leveraged by the agents in GAF-Guard. 
\subsection{Risk Atlas Nexus}
Organizations run a significant risk of legal, financial and reputational damage due of the misuse of AI systems when inadequately governed. AI governance is both an obligatory requirement and a strategic necessity.  Motivated by this, the \textit{Risk Atlas Nexus} provides open-source tooling related to governance of foundation models. It can be used to identify and prioritize risks according to the intended use-case, recommend benchmarks and risk assessments and propose mitigation strategies and concrete actions to reduce exposure to risk.  This work is described in more detail in \cite{airiskatlas2025}. The toolset has four main contributions: firstly, a centralized resource repository for a catalogue of critical governance assets such as datasets, benchmarks, taxonomies of AI risks and risk controls, which includes content ingested from IBM Risk Atlas, NIST AI RMF, MIT AI Risk Repository and more. Secondly, it provides a standards-based ontology structure to describe AI system risks and models in one coherent schema. Thirdly, structured mappings have been curated to allow linkages between heterogeneous governance resources. Finally, a python library has been created to afford developers easy integration of risk management methods - from risk identification to mitigation and control.  This structured toolkit supports the creation of more transparent and accountable AI systems, while encouraging collaboration and contributions from the broader community. By embedding LLMs into governance and risk assessment workflows, the solution advances the responsible development and deployment of AI technologies. Agents in GAF-Guard utilize the APIs from Risk Atlas Nexus for functionalities like getting risks from an intent, auto completion of questionnaires by the agent etc.  

\subsection{GAF-Guard}
\begin{figure*}
    \centering
    \includegraphics[width=0.7\linewidth]{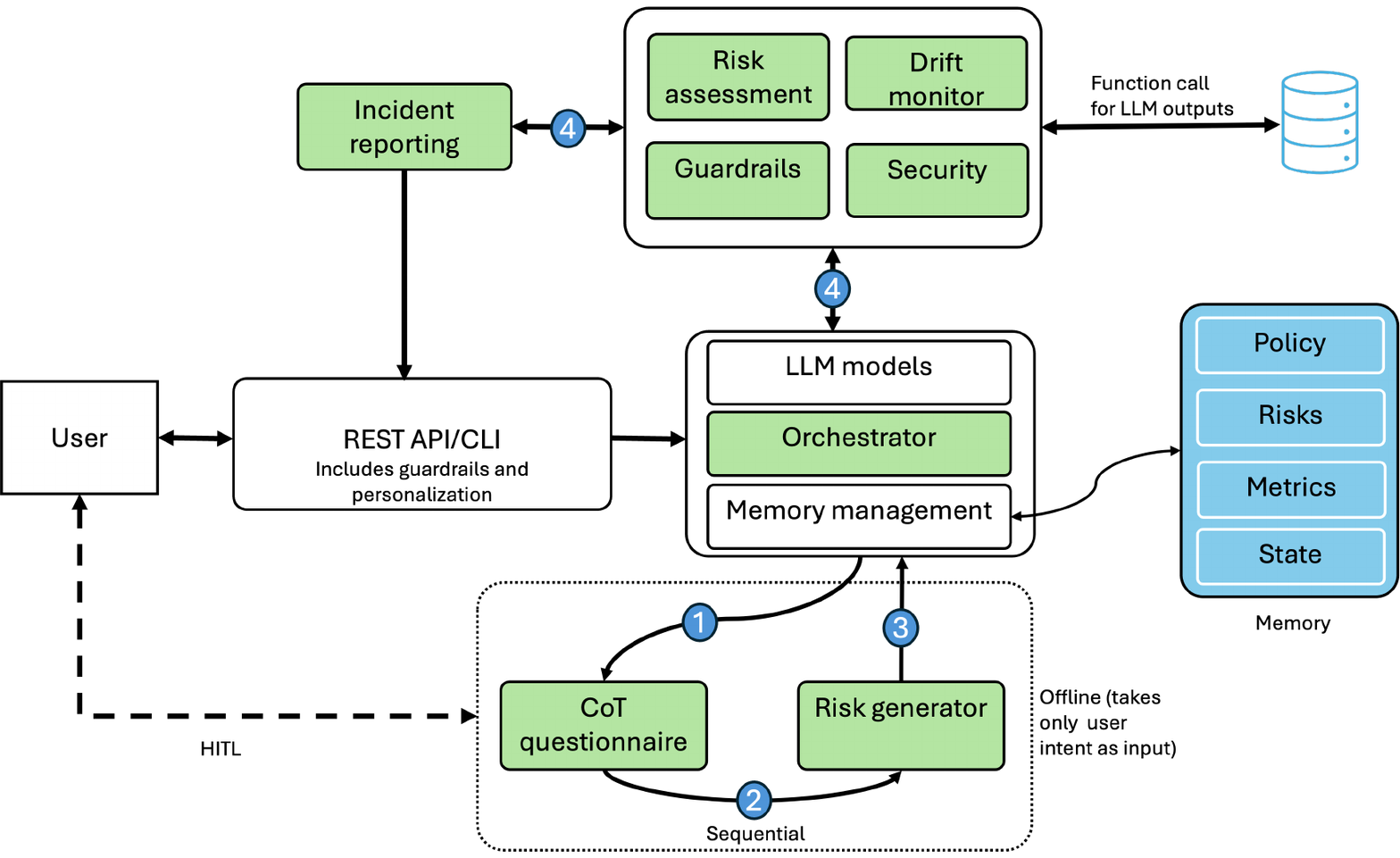}
    \caption{GAF-Guard architecture for LLM governance}
    \label{fig:arch}
\end{figure*}

We present GAF-Guard (\underline{G}overnance \underline{A}gentic \underline{F}ramework) for LLMs (Fig. \ref{fig:arch}).
This section outlines the governance framework employed for LLMs, consisting of three stages within its agentic framework:

\textbf{Pre-deployment:} The agent-based approach ensures transparent risk assessment using questionnaires, identifying potential vulnerabilities in the use-case prior to deployment.

\textbf{Post-deployment monitoring:} In the post-deployment phase, agents continuously monitor LLM outputs, detecting drifts, risk exposure, and security threats in real-time.

\textbf{Incident reporting:} An automated incident reporting system is triggered in the event of policy violations, providing immediate notification to end-users, thereby ensuring swift response and mitigation.

GAF-Guard consists of a collection of Agents addressing different capabilities as part of the governance lifecycle. These agents are orchestrated through a modular and extensible architecture that enables seamless interaction and coordination across the governance lifecycle. Each agent operates autonomously yet communicates through a centralized orchestrator, which ensures that outputs from one stage inform the next. This design supports scalability, adaptability to diverse use-cases, and integration with external governance tools or organizational policies. The framework also supports human-in-the-loop (HITL) interventions at critical junctures, reinforcing accountability and transparency in LLM governance.

\textbf{Intent to questionnaire agent: } This agent takes user input to populate a set of predefined questionnaires, providing a structured understanding of the use-case and provide the required information for downstream agent to identify the risks associated with the use-case \cite{daly2025usage}. Some of the example questions include ``What is the AI task for the given use-case'' and ``Who are the intended users of the system''. The default questionnaire for the agentic system is taken from the AIRO questionnaire \cite{airo}. To further enhance accuracy, users have the flexibility to define their own custom questions and 
provide relevant CoT examples. This approach allows for more robust risk assessment and 
mitigation compared to relying solely on zero-shot LLM responses. Function calling and JSON output ensures that the responses can be easily parsed for downstream tasks.

\textbf{Questionnaire to risk identification agent:} The user's intent, questionnaire responses, and relevant data are sequentially fed into the risk 
identification agent, enabling it to identify potential risks associated with each unique combination 
of use-case, question, and answer. The identified risks from all query scenarios are then aggregated 
and transmitted to the orchestrator for further analysis and decision-making.

\textbf{Human-in-the-Loop (HITL):} The output from the first two agents is presented to the end user for feedback, enabling them to 
provide input on either the answer choices or the final risk assessment associated with the use-case. 
This user feedback is crucial for ensuring that real-time risk monitoring is both efficient and 
consistent with the specified policy. When only the answers are revised by the user, the system 
initiates an updated query to the questionnaire-to-risk identification agent, gathering fresh risk 
assessments aligned with the revised answer choices. 

\textbf{Drift detector agents:} To enable effective real-time drift detection in user prompts, synthetic data is generated by 
leveraging the user's intent as a seed, producing relevant and irrelevant prompts of the specified use-case. These prompts serve as CoT examples, which are then used for classifying real-time user prompts for relevance using LLMs. Empirical analysis has demonstrated that utilizing CoT examples derived from the given use 
case yields substantially improved accuracy compared to generic CoT examples. Consequently, our 
approach generates synthetic data specifically tailored to the user's intent for training a CoT model 
optimized for real-time drift detection. 

\textbf{Real-time monitoring agents:} To ensure the continued efficacy and stability of the LLMs in production, 
real-time monitoring is conducted using the identified risks and user feedback-driven drift detection 
models. The Granite Guardian system is employed to detect risks in both user prompts and LLM responses, 
while custom drift detectors utilizing CoT examples from the drift detection agent 
are utilized to identify potential drifts in the deployed model.

Upon identifying a risk or detecting a drift, the incident reporting agent is triggered for further 
investigation and action. Conversely, if no risks are identified and no drifts are detected, the 
orchestrator is informed that the LLM model in production is operating within normal parameters.

\textbf{Incident reporting agent:} This agent listens to the output of the real-time drift and risk monitors and triggers a notification in case of any violations observed by the real-time monitoring agents.

\section{GAF-Guard Implementation and Evaluation}
\label{evaluation}
In this section, we will describe the implementation of the GAF-Guard framework, description of the datasets used to benchmark GAF-Guard and the evaluation methodology and metrics to measure the performance of the agents and agentic framework as a whole.

\begin{figure}
    \centering
    \includegraphics[scale=0.40]{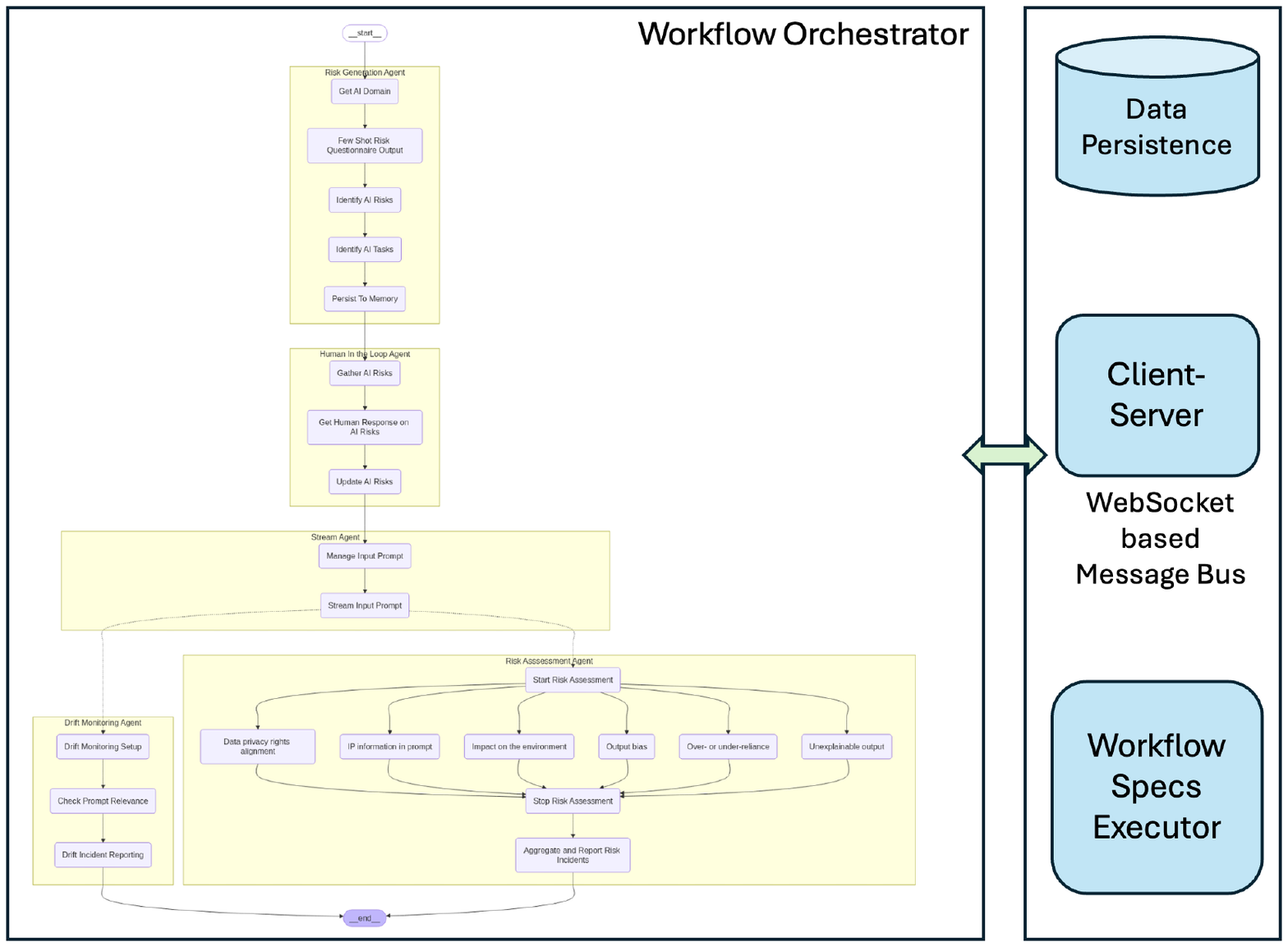}
    \caption{Graphical visualization of the architecture with langgraph implementation}
    \label{fig:langgraph}
\end{figure}

For the initial development of our agentic system, we leveraged LangChain and LangGraph that enable the creation of robust and scalable agent-based systems (Fig. \ref{fig:langgraph}). A comprehensive overview of the tools employed by the GAF-Guard framework is presented in Table \ref{tab:llm_tools}. 

Following the implementation, we conducted an evaluation at two levels: first, to assess the individual performance of the agents using standard metrics such as accuracy, precision, recall, and F1-score; second, to evaluate the agentic performance as a whole.
\begin{table}
  \caption{Tools for the agents in the governance framework} 
  \label{tab:llm_tools}
  \centering
  \begin{tabular}{ll}
    \toprule
    Agent Name         & Tool  \\ 
        \midrule
    Orchestrator & Task Delegator \\
    CoT Questionnaire   & granite3.2:8b     \\
    Risk generator   & granite3.2:8b     \\
    Risk detector  & granite3-guardian:2b     \\
    Drift monitor & llama3.2     \\
    Incident reporting   & llama3.2    \\
    Geval Evaluation & Deepseek \\
    \bottomrule
  \end{tabular}
\end{table}

\subsection{Agent evaluation}

The questionnaire agent builds upon the research in \cite{daly2025usage}, where LLMs were successfully employed to answer questionnaires. The results from this study demonstrate that a CoT approach outperforms traditional zero-shot answers for binary, dropdown, and freeform questions. By leveraging this methodology, we aim to further enhance the performance of the questionnaire and risk extraction agents by incorporating a Human-In-The-Loop component in the end, which enables us to incorporate user feedback and refine our responses and final risks of the use-case accordingly.

Based on the questionnaire, the domain of the business use-case is identified and this information is used for creating the drift monitors which can assess in realtime whether the prompts are related to the domain.

\begin{figure*}
    \centering
    \includegraphics[width=0.5\linewidth]{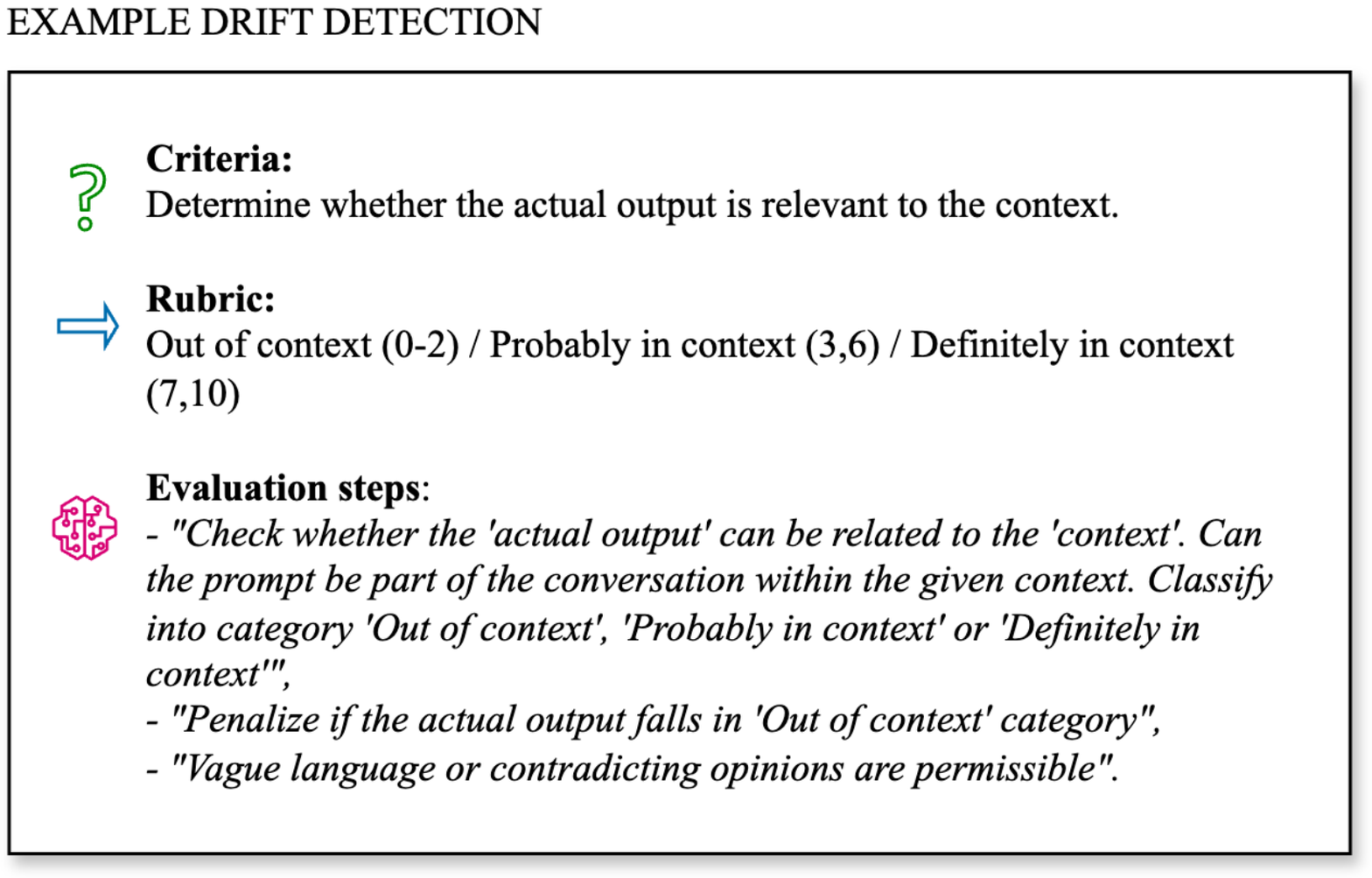}
    \caption{Drift detection using G-Eval}
    \label{fig:example_drift_detection}
\end{figure*}

We employ llm-as-a-judge paradigm, G-Eval, and Granite Guardian as a tool for detecting drift and identifying risks respectively. The G-Eval criteria for detecting drift, developed using \cite{Ip_deepeval_2025}, are shown in Fig. \ref{fig:example_drift_detection}. By utilizing standard classification metrics such as precision, recall, accuracy, and F1 score, we assess the objective outcomes of the drift detection task.

For the initial phase, Drift Monitor's development focused on identifying prompt-drift issues, specifically those where user-input prompts stray from the context. Four distinct approaches were developed and subsequently evaluated and compared to determine an effective method for detecting such deviations: 
\begin{itemize}
    \item Employing G-Eval and relevance to context as the primary metric for evaluating correctness. The metric score obtained is consolidated with the historical values through rolling average of drift. This rolling average is compared to a user-defined threshold to identify instances of drift in prompts. 
    \item Generating static CoT examples and leveraging these for classification purposes.
    \item Dynamic synthetic CoT examples based on the context are created and these examples are coupled with an LLM for classification.
    \item Zero-shot approach which involved directly querying a LLM regarding the relevance of the prompt within the context.
\end{itemize}

\begin{figure*}
    \centering
    \includegraphics[width=.95\linewidth]{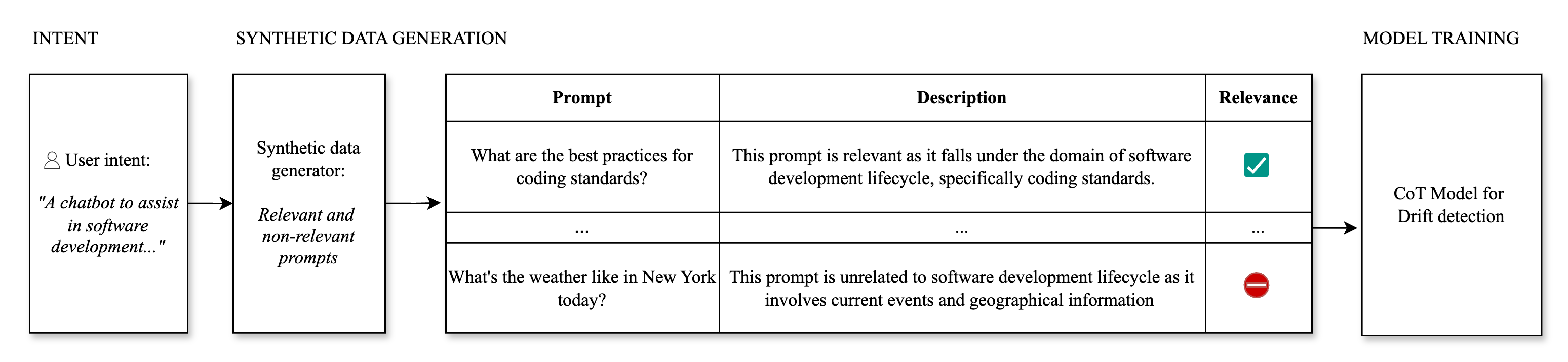}
    \caption{Sample synthetic data from software development domain to test the performance of the drift detector.}
    \label{fig:drift_detect_1}
\end{figure*}

 To assess the effectiveness of the drift monitor, synthetic data was created for evaluation purposes. User prompts were crafted for the software development domain. Of these, some prompts accurately reflected domain-specific scenarios, while remaining were generated outside the domain to simulate real-world challenges. A sample of the synthetic data for software development are shown in Fig. \ref{fig:drift_detect_1}.

The results of all models are presented in Table \ref{tab:classification_metrics} for the software development use-case. It can be seen that G-eval based drift monitor exhibits superior performance in drift detection task. Dynamic synthetic CoT example generation yields comparable results, although it requires the agent to generate novel examples for each domain, which may introduce context-dependent variability and quality of the synthetic data generated. Zero shot approach is not shown in the table since the LLM classifies all prompts as relevant to the domain. Both, G-eval based drift monitor and dynamic synthetic CoT drift monitors are included in GAF-Guard. 
\begin{table}
  \caption{Classification metrics for drift monitor for software development domain.}
  \label{tab:classification_metrics}
  \centering
  \begin{tabular}{lllll}
    \toprule
    Drift method     & Accuracy     & Precision  & Recall & F1   \\ 
        \midrule
    G-Eval & 0.86 & 0.87 & 0.86 & 0.86     \\
    Static CoT & 0.66 & 0.72 & 0.66 & 0.66    \\
    Dynamic CoT & 0.83 & 0.87 & 0.83 & 0.81    \\
    \bottomrule
  \end{tabular}
\end{table}

Granite Guardian \cite{padhi2024graniteguardian} is a fine-tuned Granite language model to help with risk detection across multiple risk dimensions like harm, answer relevance etc. Granite Guardian has been benchmarked in \cite{padhi2024graniteguardian} on multiple benchmark datasets like AegisSafetyTest \cite{ghosh2024aegis}, HarmBench Prompt \cite{mazeika2024harmbench} etc and against multiple baselines like Llama-Guard \cite{inan2023llama} and ShieldGemma \cite{zeng2024shieldgemma} series of models and demonstrate competitive performance.

\subsection{Agentic framework evaluation with $\tau$-bench}
To comprehensively evaluate the agentic performance, we leveraged the $\tau$-bench evalutaion tool \cite{yao2024tau}, which can simulate dynamic user-agent interactions across multiple trials. The $\tau$-bench metric (k\textasciicircum n) was utilized to evaluate both the accuracy of individual responses and the overall reliability and robustness of the agents in managing consistent responses. 

\begin{figure*}
    \centering
    \includegraphics[scale=0.40]{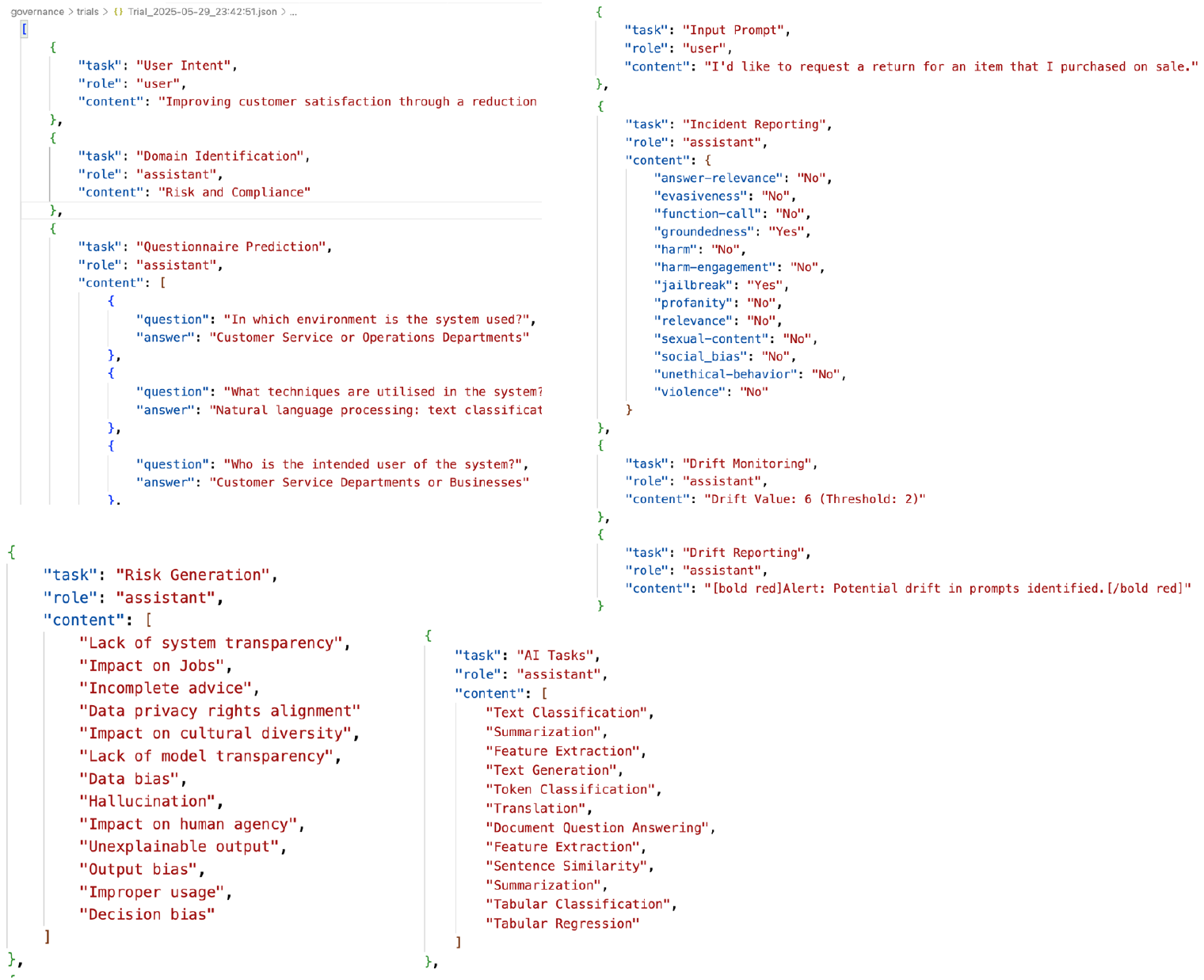}
    \caption{Snippets of a JSON file used in benchmarking Agentic Workflow}
    \label{fig:benchmark_json}
\end{figure*}

Further, we created a synthetic dataset as a JSON file for benchmarking it using $\tau$-bench \cite{yao2024tau}. 
The JSON file details the trajectories of agents during the workflow execution phase. Each run of an agent is recorded as a trajectory by the system. An example trajectory, presented in the form of a JSON file, is shown in Fig. \ref{fig:benchmark_json}. Each trajectory is represented as a JSON list, where each entry is a dictionary detailing a single task completed by a worker agent. These tasks are recorded in the order they are scheduled for execution by the agent. Each dictionary includes the task name, the role (either user or assistant), and the content that displays the agent's actual response. For benchmarking purposes, we compare each captured trajectory against a manually prepared ground truth trajectory. We utilized the G-Eval metric \cite{liu2023g}, which uses LLM-as-a-judge with chain-of-thoughts (CoT) to assess LLM outputs and assign rewards. The results of this comparison are then processed by the $\tau$-bench function to derive the final score. Our agentic workflow dataset includes 10 trajectories of user and agent responses in the order of execution, which helps evaluate the agents' responses to user queries.

The agentic governance workflow we proposed utilizes an orchestrator-worker model, where the orchestrator breaks down a task and assigns each sub-task to different worker agents. It supports scalable multi-agent systems, with each agent assigned a specific task. There are four worker agents involved: Risk Generation, Human in the Loop (HITL), Risk Assessment, and Drift Monitoring. The orchestrator is responsible for delegating tasks to these worker agents, managing data serialization and memory persistence, aggregating the results, and sending them to the caller. We utilized a $\tau$-bench (pass\textasciicircum k) metric to assess the reliability of agent behaviour across multiple trials. Our experiments (Table \ref{tab:tau-bench-scores}) demonstrate that our agentic system achieved an average accuracy of approximately 90\%, and maintained a high level of consistency, even after three runs (pass\textasciicircum 3 > 80\%) in customer complaints use-cases). This highlights the capability of the agent workflow to perform tasks consistently and execute them reliably.

\begin{table}
  \caption{$\tau$-bench scores for the Governance Workflow}
  \label{tab:tau-bench-scores}
  \centering
  \begin{tabular}{lllll}
    \toprule
    Use-case     & Average     & k\textasciicircum 1  & k\textasciicircum 2 & k\textasciicircum 3   \\ 
        \midrule
    Customer complaints & 0.96 & 0.92 & 0.88 & 0.87     \\
    \bottomrule
  \end{tabular}
\end{table}


\section{Conclusion and Future work}
\label{conclusion}
We present an agentic governance framework that has been designed and implemented to address the risk management needs of LLM applications. By leveraging minimal human interaction and modular agent integration, the framework enables efficient and adaptive risk detection and mitigation. The flexibility of the design allows users to tailor the framework to their specific requirements if required, ensuring a bespoke governance structure that aligns with their unique user intent and preferences. By taking into account the specific needs and goals of each end-user, the agentic governance framework is better positioned to provide a tailored approach to risk management, ultimately leading to a more effective and contextually relevant governance structure for real world LLM implementations. Future work will consider policy integration into the agentic framework. Also, while the system can provide information on risks associated with a use-case, the value of these risk identification and detection tasks can be enhanced if their severity is accurately assessed which will also be developed in the future. 

\newpage

\bibliographystyle{unsrt}  
\bibliography{references}  

\appendix
\section{Limitations}
\label{limitations}
The paper presents an innovative agentic system for governance, offering substantial advantages in identifying and managing risks. However, there are some limitations inherent in the current framework whichwe will try to address in the future work:
\begin{itemize}
    \item Despite being able to identify risks associated with a use-case, the system does not provide theoretical guarantees of detecting all potential risks. This can potentially be an issue if the agentic framework is deployed without human feedback.
    \item The agentic system's ability to deploy dedicated monitors for mitigators or guardrails is an essential consideration, especially when mitigation measures are included in the deployment. Exclusive monitors for mitigators would enhance the robustness of the framework by providing real-time performance monitoring, thereby reflecting the effectiveness of risk mitigation strategies.
    \item In the current investigation, LLMs with fixed architectures were employed for the experiments reported herein. While the framework was designed to be adaptable to various LLMs, a comprehensive analysis of its generalized performance is still lacking. Future work will focus on conducting an extensive ablation study to thoroughly investigate the framework's robustness and versatility across different LLM variants.
\end{itemize}
The limitations mentioned above can be addressed by integrating pre-defined and dynamic user preferences. The framework provides the initial groundwork for the community to contribute any additional features or address the limitations like the ones mentioned in this section.

\end{document}